# Using Qualitative Relationships for Bounding Probability Distributions

**Chao-Lin Liu and Michael P. Wellman**
University of Michigan AI Laboratory
Ann Arbor, Michigan 48109, USA
{chaolin,wellman}@umich.edu

## Abstract

We exploit qualitative probabilistic relationships among variables for computing bounds of conditional probability distributions of interest in Bayesian networks. Using the signs of qualitative relationships, we can implement abstraction operations that are guaranteed to bound the distributions of interest in the desired direction. By evaluating incrementally improved approximate networks, our algorithm obtains monotonically tightening bounds that converge to exact distributions. For supermodular utility functions, the tightening bounds monotonically reduce the set of admissible decision alternatives as well.

## 1 Introduction

Approximation techniques have gained increasing interest among those employing Bayesian networks for probabilistic reasoning, despite the fact that computing a desired probability distribution to a fixed degree of accuracy has been shown to be NP-hard (Dagum & Luby 1993). Approximation techniques offer reasonable prospects of significant accuracy, and increased opportunity to consider applications larger than we could otherwise. For instance, approximation techniques can be useful for applications that need to respond to requests for solutions under time constraints. By appropriately managing the reasoning process, we may obtain approximate solutions that meet the needs of these applications in cases where we would not be able to compute exact solutions given the time constraints.

Researchers have explored a wide range of techniques to enable evaluation algorithms to compute approximations of desired probability distributions. One common approach is to compute point-valued approximations of desired distributions. For instance, stochastic simulation algorithms compute approximations of desired distributions with random numbers sampled based on the given Bayesian network (Pearl 1987; Henrion 1988; Neal 1993). Some other algorithms compute approximations by ignoring information that specifies the exact distribution in the given network, for example, state-space abstraction (Wellman & Liu 1994) and arc removal (van Engelen 1997).

Another popular approach is to compute bounds or intervals of the desired probability distributions. For instance, *bounded conditioning* computes bounds of probabilities by limiting the number of cutset instances used in computation (Horvitz, Suermondt, & Cooper 1989). *Localized partial evaluation* computes intervals of probability values by ignoring selected nodes in the network (Draper & Hanks 1994).

One advantage of computing probability intervals over point-valued approximations is that intervals explicitly specify a range that contains exact solutions, thereby providing information about bounds of possible errors. Algorithms that compute point-valued approximations typically do not provide comparable information. Also, many algorithms that compute probability intervals can control the tightness of the intervals by tuning the amount of neglected information, thereby improving bounds when allocated more computation time. As a result, when used in anytime computation (Horvitz 1990; Boddy & Dean 1994), these algorithms guarantee monotonic improvement of the quality of returned solutions for any problem instance. In contrast, the algorithms that compute point-valued approximations usually expect the approximations to improve with the allocated computation time only on average.

In this paper, we extend our iterative state-space abstraction (ISSA) algorithm (Wellman & Liu 1994) to compute the bounds of cumulative distribution functions (CDFs) of interest. The extended algorithm takes advantage of qualitative relationships among variables in the computation. The qualitative relationships that summarize special quantitative dependence relationships among variables are as originally defined for qualitative probabilistic networks (QPNs) (Wellman 1990). When variables in Bayesian networks exhibit these special quantitative relationships, it is possible to compute bounds of conditional CDFs of interest using such relationships. We report conditions under which the extended ISSA algorithm can compute bounds of conditional CDFs, and show that these bounds tighten monotonically with iterations. When used with supermodular utility functions, the tightening bounds imply reduction of the set of decision alternatives that might be optimal, thereby helping decision makers to focus on fewer decision alternatives.

Next, we review definitions of qualitative relationships and define bounds of cumulative distribution functions. Then, we discuss conditions for computing bounds of CDFs, and how to control the tightness of these bounds. In Section 4, we review our ISSA algorithm and discuss extensions of the algorithm for computing bounds. In Section 5, we describe some applications of the bounds, and, in Section 6, we compare and contrast our algorithm with existing algorithms designed for computing probability intervals.

## 2 Background

### 2.1 Qualitative relationships

The qualitative relationships we employ are based on those defined for qualitative probabilistic networks (QPNs) (Wellman 1990). QPNs are abstractions of Bayesian networks, with conditional probability tables summarized by the signs of qualitative relationships between variables. Each arc in the network is marked with a sign—positive (+), negative (−), or ambiguous (?)—denoting the sign of the qualitative probabilistic relationship between its terminal nodes.

The interpretation of such qualitative influences is based on *first-order stochastic dominance* ($FSD$) (Fishburn & Vickson 1978). Let $F(x)$ and $F'(x)$ denote two CDFs of a random variable $X$. Then $F(x)\ FSD\ F'(x)$ holds if and only if (iff) $F(x) \leq F'(x)$ for all $x$. We say that one node positively influences another iff the latter's conditional distribution is increasing in the former, all else equal, in the sense of $FSD$.

**Definition 1 ((Wellman 1990))** *Let $F(z|x_i, y)$ be the cumulative distribution function of $Z$ given $X = x_i$ and the rest of $Z$'s parents $Y = y$. Node $X$ positively influences node $Z$, denoted $S^+(X, Z)$, iff, for all $x_i, x_j, y$,*

$$x_i \leq x_j \Rightarrow F(z|x_j, y)\ FSD\ F(z|x_i, y).$$

Analogously, we say that node $X$ *negatively influences* node $Z$, denoted $S^-(X, Z)$, when we reverse the direction of the dominance relationship in Definition 1. The arc from $X$ to $Z$ in that case carries a negative sign. When the dominance relationship holds for both directions, we denote the situation by $S^0(X, Z)$. However, this entails conditional independence, and so we typically do not have a direct arc from $X$ to $Z$ in this case. When none of the preceding relationships between the two CDFs hold, we put a question mark on the arc, and denote such situations as $S^?(X, Z)$.

We may apply the preceding definitions to boolean nodes under the convention that **true** > **false**.

In this paper, we extend the notation to express conditional influences, and use it in a slightly more general way. The expression $S^+(X, Z\|c)$ means that $F(z|x, c)$ is decreasing in $x$ given $C = c$, and $S^-(X, Z\|c)$ means that $F(z|x, c)$ is increasing in $x$ given $C = c$. Moreover, $X$ and $C$ do not have to be parents of $Z$.

### 2.2 Bounds of probability distributions

**Definition 2** *A CDF $\overline{F}(x)$ is an upper bound of $F(x)$, if $F(x)\ FSD\ \overline{F}(x)$. A CDF $\underline{F}(x)$ is a lower bound of $F(x)$, if $\underline{F}(x)\ FSD\ F(x)$.*

Notice that lower and upper probabilities (Chrisman 1995) and bounds of CDFs are related concepts. With the bounds of CDFs, we may define lower and upper probabilities of $X$. Let $M$ denote the event that $x_i < X \leq x_j$. The lower and upper probabilities of $M$, denoted by $\underline{\Pr}(M)$ and $\overline{\Pr}(M)$, respectively, are:

$$\underline{\Pr}(M) = \max(0, \underline{F}(x_j) - \overline{F}(x_i)),\ \text{and}$$
$$\overline{\Pr}(M) = \overline{F}(x_j) - \underline{F}(x_i).$$

$\overline{\Pr}(M)$ is guaranteed to be between 0 and 1 since $1 \geq \overline{F}(x_j) \geq \overline{F}(x_i) \geq \underline{F}(x_i) \geq 0$.

## 3 Bounding probability distributions

In this section, we report methods for computing bounds of selected conditional CDFs in Bayesian networks. These methods take advantage of qualitative relationships among nodes for bounding probability distributions. We also present methods for controlling the tightness of bounds.

We can compute bounds of some probability distributions by strengthening and weakening selected CDFs in Bayesian networks. Let $Y$ be a child of $A$, and denote the set of parents of $Y$ excluding $A$ by $PX(Y)$. For simplicity henceforth, we denote $F(Y = y|A = a, PX(Y) = px(Y))$ by $F(y|a, px(Y))$. We *strengthen* $F(y|a, px(Y))$ with respect to $A$ by replacing $F(y|a, px(Y))$ with $F'(y|a, px(Y))$ such that

$$F'(y|a, px(Y))\ FSD\ F(y|a, px(Y)),\ \text{for all}\ a.$$

The most important effect of strengthening the CDF $F(y|a, px(Y))$ with respect to $A$ is to increase the probability of $Y$ being a larger value for some states of $A$. Analogously, we *weaken* CDF $F(y|a, px(Y))$ when the FSD relationship is reversed, and weakening $F(y|a, px(Y))$ with respect to $A$ implies that we decrease the probability of $Y$ being a larger value for some states of $A$.

Using these strengthening and weakening operations, we may compute the bounds of selected conditional probabil-ity distributions when some of the links of a Bayesian network can be marked with decisive qualitative signs: "+" or "−". Consider the network in Figure 1, and assume that we strengthen $F(y|x)$ with respect to $X$ by replacing $F(y|x)$ with $F'(y|x)$ such that $F'(y|x)$ $FSD$ $F(y|x)$ for all states of $X$. Given that $S^+(Y, Z)$ implies that $F(z|y)$ is decreasing in $y$, our strengthening $F(y|x)$ will decrease $F(z|x)$, since the probability of $Y$ being larger has been increased for all states of $X$. Therefore, we have obtained a lower bound of $F(z|x)$ for all $x$ by strengthening $F(y|x)$. As a result, we also have obtained a lower bound of $F(z|e)$ for all $e$ because $F(z|e) = \sum_X F(z|x)\Pr(x|e)$.

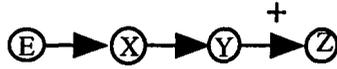

Figure 1: A Bayesian network with a link marked with a qualitative sign.

This example illustrates that we may compute bounds of CDFs by *locally* strengthening selected CDFs. Specifically, given $S^+(Y, Z)$, we are able to compute a lower bound of $F(z|e)$ by strengthening $F(y|x)$ with respect to $X$. The strengthening of $F(y|x)$ can be carried out by using the values in the conditional probability tables (CPT) associated with $Y$.

In the following theorem statements, we use *ancestral ordering* of nodes as defined below.

**Definition 3 (cf. (Neapolitan 1990))** *Let $J$ denote a set of nodes $\{J^1, \ldots, J^n\}$ in a Bayesian network. $[J^1, \ldots, J^n]$ is an ancestral ordering of the nodes in $J$ if for every $J^i \in J$ all the ancestors of $J^i$ are ordered before $J^i$.*

The following theorem presents conditions for computing bounds of a conditional CDF of a variable $Z$ given the evidence $E = e$ by strengthening and weakening the distributions of the children of a distinguished node $A$. We call nodes whose values are instantiated *evidence nodes*, and we denote the set of evidence nodes by $E$. Let $Y$ be the children of $A$ and $Y^i$ be a node in $Y$. The theorem is applicable when children of $A$ meets the stated requirements. We denote the subset $Y \setminus \{Y^i\}$ of $Y$ by $SB(Y^i)$, and we use the notation $S^{\sigma^i}(Y^i, Z\|e, X)$ to represent that $S^{\sigma^i}(Y^i, Z)$ given $E = e$ and all possible instantiations of $X$, where $\sigma^i$ is a sign for qualitative relationship between $Y^i$ and $Z$.

**Theorem 1** *Assume that:*

1. *For all $i$, $S^{\sigma^i}(Y^i, Z\|e, SB(Y^i))$, where $\sigma^i$ is either +, −, or 0.*

2. $CI(Z, \{E, Y\}, A)$.

3. *$E, A,$ and $Y$ appear in order in an ancestral ordering.*

4. *$Y^i$ is not a descendant of nodes in $SB(Y^i)$.*

*When $\sigma^i = -$, we obtain, respectively, a lower and an upper bound of $F(z|e)$ by weakening and strengthening $F(y^i|a, \boldsymbol{px}(Y^i))$ with respect to $A$. When $\sigma^i = +$, we obtain, respectively, an upper and a lower bound of $F(z|e)$ by weakening and strengthening $F(y^i|a, \boldsymbol{px}(Y^i))$ with respect to $A$. When $\sigma^i = 0$, neither strengthening nor weakening $F(y^i|a, \boldsymbol{px}(Y^i))$ with respect to $A$ affects $F(z|e)$.*

Proof. *Proof for this theorem is an extension of our explanation for the preceding example where node $X$ is $A$ in the theorem. This theorem and its proof extend the basic ideas to consider the case when $A$ has multiple child nodes. (Due to space limitations, proofs are available only in the full paper, available at http://www.umich.edu/~chaolin/.)*

**Example 1** *In the network shown below, we have (a) $S^+(Y1, Z\|e, Y2)$ and $S^-(Y2, Z\|e, Y1)$ for any $e$, (b) $CI(Z, \{E, Y1, Y2\}, A)$, (c) $[E, A, Y1, Y2]$ is an ancestral ordering, and (d) $Y1$ is not a descendant of $Y2$ and vice versa. Therefore, we can obtain a lower bound of $F(z|e)$ for any $E = e$ by strengthening $F(y1|a)$ or weakening $F(y2|a)$ with respect to $A$.*

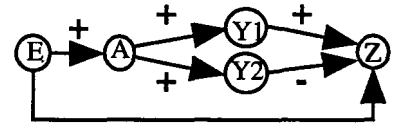

Figure 2: An applicable example for Theorem 1.

Theorem 1 can be applied to cases where we strengthen and weaken multiple such $F(y^i|a, \boldsymbol{px}(Y^i))$, as long as these strengthening and weakening operations are coordinated consistently so that the effects are to find a lower or an upper bound of $F(z|e)$. This extended interpretation of the theorem can be proved inductively as follows. The theorem dictates that we obtain a bound of the exact $F(z|e)$ by strengthening or weakening one particular $F(y^i|a, \boldsymbol{px}(Y^i))$. In addition, we can obtain a new bound of $F(z|e)$ by strengthening or weakening one more such $F(y^i|a, \boldsymbol{px}(Y^i))$ in the ABN where $(n-1)$ $F(y^i|a, \boldsymbol{px}(Y^i))$ has been strengthened or weakened. Therefore, by induction, we may coordinate the strengthening and weakening operations to obtain lower (or upper) bounds of $F(z|e)$ by strengthening or weakening the conditional probability distributions of all nodes in $Y$ with respect to $A$.

The first condition of Theorem 1 requires that $Y^i$ have a non-ambiguous qualitative relationship with $Z$. This qualitative relationship determines the selection of strengthening and weakening operations for computing desired bounds. The remaining conditions ensure that we can compute desired bounds by locally modifying $F(y^i|a, \boldsymbol{px}(Y^i))$. Specifically, when $A$ has multiple child nodes $Y$, we can, simultaneously and independently, strengthen or weaken the conditional probability distribution of each node in $Y$ to obtained bounds of $F(z|e)$. Notice that this and the following theorem do not require a decisive qualitative relationship between the evidence nodes $E$ and the node of interest $Z$.

**Example 2** *Theorem 1 may be applicable to networks that are as complex as the one shown below. In this network, we assume all links point from the left to the right hand side, and we use thick gray links to represent bunches of links that might exist between clouds of nodes and individual nodes. With $E = \{E1, E2, E3\}$, we can verify that $CI(Z, \{E, Y\}, A)$ holds in this network. Also, $[E, A, Y]$ is an ancestral ordering, and $Y^i$ is not a descendant of $SB(Y^i)$ for all $Y^i$. Therefore, Theorem 1 is applicable to any $Y^i$ that satisfies the first condition in the theorem.*

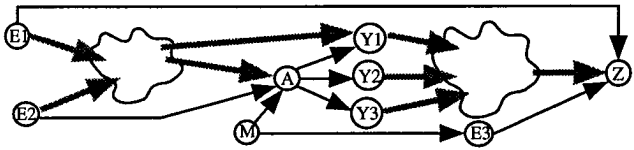

Figure 3: Another applicable example for Theorem 1.

Theorem 1 cannot be applied to cases where $A$ is a parent node of $Z$ because $Y$ and $Z$ represent distinct nodes. The following theorem specifies conditions and methods for abstracting the parents of $Z$ to compute bounds of $F(z|e)$.

**Theorem 2** *In addition to Conditions 3 and 4 of Theorem 1, assume that $Z \in Y$. We obtain, respectively, a lower and an upper bound of $F(z|e)$ by strengthening and weakening $F(z|a, px(Z))$ with respect to $A$.*

**Example 3** *In the network shown in Figure 2, we have (a) [E,Y1,Z] is an ancestral ordering and (b) Z is the only descendant of Y1. Therefore, Theorem 2 is applicable, and we can obtain a lower bound of $F(z|e)$ by strengthening $F(z|y1)$ with respect to Y1. Analogously, we obtain a lower bound of $F(z|e)$ by strengthening $F(z|y2)$ with respect to Y2.*

Theorems 1 and 2 also provide guidelines for obtaining tighter bounds. For convenience, if $H(x)$ $FSD$ $G(x)$ $FSD$ $F(x)$, we say that $G(x)$ is less dominating than $H(x)$. Roughly speaking, the following corollary, which follows from Theorem 1, states that we can obtain tighter bounds of $F(z|e)$ by setting $F(y^i|a, px(Y^i))$ to a less (or more) dominating alternative.

**Corollary 1** *Let $G(y^i|a, px(Y^i))$ and $H(y^i|a, px(Y^i))$ be alternatives for weakening $F(y^i|a, px(Y^i))$ with respect to $A$. Assume that $H(y^i|a, px(Y^i))$ is less dominating than $G(y^i|a, px(Y^i))$ for all $a$ and $px(Y^i)$. Then, weakening $F(y^i|a, px(Y^i))$ by $G(y^i|a, px(Y^i))$ rather than $H(y^i|a, px(Y^i))$ provides a tighter lower bound of $F(z|e)$ when $\sigma^i = -$, and a tighter upper bound of $F(z|e)$ when $\sigma^i = +$.*

*Analogously, strengthening $F(y^i|a, px(Y^i))$ by a less dominating CDF provides a tighter upper bound of $F(z|e)$ when $\sigma^i = -$, and a tighter lower bound of $F(z|e)$ when $\sigma^i = +$.*

Similarly, we may derive the following corollary from Theorem 2. This corollary provides guidelines for obtaining tighter bounds in applying Theorem 2.

**Corollary 2** *Applying Theorem 2, we obtain tighter lower (upper) bounds of $F(z|e)$ by setting $F(z|a, px(Z))$ to a less (more) dominating alternative.*

Notice that neither Theorem 1 nor Theorem 2 requires any particular qualitative relationship between $A$ and nodes in $Y$. As we mention in Section 4.1, the existence of a particular qualitative relationship between $A$ and nodes in $Y$ facilitates, but is not required for, the application of the theorems.

## 4 State-space abstraction

In previous work, we report an iterative state-space abstraction (ISSA) algorithm for approximate evaluation of Bayesian networks (Wellman & Liu 1994). The ISSA algorithm aggregates states of variables into *superstates* to construct abstract versions of the *original Bayesian networks* (OBNs) that specify exact probability distributions. We use these abstract Bayesian networks (ABNs) to compute point-valued approximations of the probability distributions of interest.

To construct ABNs, we select some nodes, called *abstracted nodes*, from the OBNs, and aggregate their states. As a result of state aggregation, we need to assign the CPTs of both the abstracted nodes and their child nodes. We call the method used in this assignment task a CPT assignment *policy*. In this section, we introduce the *dominance policy* for computing bounds of CDFs.

### 4.1 Dominance policy

Recall that we need to assign the CPTs of the abstracted node and its child nodes when we abstract a node. The *dominance policy* modifies the CPT of the abstracted node $A$ as follows:

$$\hat{\Pr}([a_{k,l}]|pa(A)) = \sum_{j=k}^{l} \Pr(a_j|pa(A)), \quad (1)$$

where $PA(A)$ is the set of parents of $A$, and $[a_{k,l}]$ is the superstate representing the aggregation of states from $a_k$ through $a_l$, $k \leq l$.

We may choose to strengthen or weaken selected conditional probability distributions, depending on whether we want to compute lower or upper bounds of the desired CDFs. Let $Y$ be a child node of $A$, and $PX(Y)$ be the subset of parent nodes of $Y$ excluding $A$. If we choose to *strengthen* the conditional probability distributions of $Y$ with respect to $A$, we assign the CPT of $Y$ as follows.

$$\hat{F}(y|[a_{k,l}], px(Y)) = \min_{j \in [k,l]} F(y|a_j, px(Y)) \quad (2)$$

If we choose to *weaken* the conditional probability distributions of $Y$ with respect to $A$, we assign the CPT of $Y$ as follows.

$$\hat{F}(y|[a_{k,l}], px(Y)) = \max_{j \in [k,l]} F(y|a_j, px(Y)) \quad (3)$$

We need to compare the probability values of the states $a_j$, $j = k, l$, aggregated in a superstate $[a_{k,l}]$ in the OBNs and ABNs in order to apply the aforementioned theorems in analyzing the effects of the dominance policy. In terms of Theorem 1, we want to strengthen or weaken $F(y^i|a, px(Y^i))$ to obtain desired bounds when we abstract node $A$. However, $A$ in an ABN has fewer states than $A$ in the OBN, so for a superstate $[a_{k,l}]$, we do not have corresponding $\hat{\Pr}(y^i|a_j, px(Y^i))$ for all $j = k, l$ in the ABN. Fortunately, we may show that the strengthening and weakening operations in the dominance policy have the effects of strengthening and weakening distributions that we defined at the beginning of Section 3. We show this by transforming the ABNs into equivalent networks and comparing these equivalent networks with the OBNs. The procedure for constructing equivalent networks of ABNs and related proofs are available in the full paper.

Application of (2) and (3) becomes easier when the links from the abstracted node to its child nodes can be marked with decisive qualitative signs. For instance, if the sign from $A$ to $Y$ is "+", the right hand sides of (2) and (3) are simply $F(y|a_l, px(Y))$ and $F(y|a_k, px(Y))$, respectively. However, the application of (2) and (3) does not require any particular qualitative relationship between $A$ and $Y$.

### 4.2  Single node abstraction

In this section, we report the application of the ISSA algorithm to computing bounds of distributions using qualitative relationships among nodes. We discuss the application of Theorems 1 and Corollary 1. The application of Theorem 2 and Corollary 2 is analogous.

We operationalize Theorem 1 using the state-space abstraction methods. To compute bounds of the desired CDFs $F(z|e)$, we can abstract the state space of any node $A$ that meets the conditions of the theorem. We may apply the inference algorithms for QPNs (Druzdzel & Henrion 1993) to locate those nodes whose children have unambiguous qualitative relationships with $Z$ as specified in the first condition of the theorem. We apply (1) to assign the CPT of $A$, and we apply (2) or (3) to assign the CPTs of the child nodes of $A$. The selection of (2) or (3) depends on whether we want to compute lower or upper bounds of the desired CDFs, and Theorem 1 provides guidelines for the selection.

The CDF $F(z|e)$ specified in the ABNs constructed in this manner is a bound of the CDFs of the $F(z|e)$ specified in the OBNs. This is due to Theorem 1 and the fact that we can show that the effects of applying the dominance policy in abstracting nodes are equivalent to strengthening (or weakening) the conditional probability distributions of the children of the abstracted nodes.

We can show that ISSA returns bounds that tighten in each iteration using Corollary 1. The tightening bounds are due to the fact that, as we split superstates, the reassigned CDFs, respectively, become less and more dominating when we strengthen and weaken the original CDFs. Consider the case in which we want to strengthen $F(y|a, px(Y))$ with respective to $A$. When we split the superstate $[a_{k,l}]$ into two superstates $[a_{k,m}]$ and $[a_{m+1,l}]$, $m \in (k,l)$, $F(y|[a_{k,l}], px(y))$ is replaced with $F(y|[a_{k,m}], px(y))$ and $F(y|[a_{m+1,l}], px(y))$. We can easily verify that $F(y|[a_{k,l}], px(y))$ is not larger than $F(y|[a_{k,m}], px(y))$ and $F(y|[a_{m+1,l}], px(y))$ with (2). As a result, the newly reassigned CDFs are less dominating, and, according to Corollary 1, the bounds of $F(z|e)$ tighten in each iteration of ISSA.

### 4.3  Multiple node abstraction

We may compute bounds of CDFs by abstracting multiple nodes that do not share child nodes. With an analogous method used in the previous section, we can show that the bounds obtained by evaluating network with multiple abstracted nodes tighten as we split superstates.

Without loss of generality, we may assume that the purpose of abstracting nodes is to compute an upper bound of $F(z|e)$. We analyze the effects of abstracting multiple nodes by assuming that we abstract one node at a time and that we abstract nodes $A^i$, $i = 1, 2, \ldots, m$. Let $ABN^i$ denotes the ABN that is constructed by sequentially abstracting node $A^1$ up to node $A^i$. Clearly, by applying the result from the previous section, the $F(z|e)$ specified in $ABN^i$ is an upper bound of the $F(z|e)$ specified in $ABN^{i-1}$ since one more node is abstracted in $ABN^i$ than in $ABN^{i-1}$. Therefore, by induction, we can show that the $F(z|e)$ specified in $ABN^m$ is an upper bound of the $F(z|e)$ specified in the OBN.

The remaining problem is to show that the $ABN^m$ that is constructed by sequentially abstracting node $A^1$ through $A^m$ is the same as the ABN that is constructed by simultaneously abstracting all $A^i$'s. Recall that we use maximal and minimal operations for strengthening and weakening CDFs in (2) and (3). As a result, the order that we abstract the nodes matters only when the abstracted nodes share child nodes. When abstracted nodes share child nodes, abstracting nodes in different orders may result in different abstract networks due to the fact that maximal and minimal operations are not commutative. However, when the abstracted nodes do not share child nodes as in our case, the ordering of these nodes being abstracted will not affect the resulting network, and the $ABN^m$ that is constructed by sequentially abstracting node $A^1$ through $A^m$ is the same as the ABN that is constructed by abstracting all $A^i$'s in any order. Therefore, we have shown that we can abstract multiple nodes that do not share child nodes to obtain bounds of CDFs.

### 4.4 Generalized qualitative relationship

We can extend the applicability of the theorems by relaxing the required qualitative relationship. Consider the CDF $F(y|x, c)$, where $c$ denotes instantiations of conditioning variables other than $x$. Assume that $X$ has $m$ states, $x_1, x_2, \ldots, x_m$. We use $S_n^+(X, Y\|c)$ to denote the situation in which $F(y|x, c)$ has the following property: $n$ is the smallest integer such that, for any $y$, $i \in [1, m], j \in [n, m), i + j \leq m$,

$$F(y|x_{i+j}, c) \ FSD \ F(y|x_i, c).$$

Analogously, we use $S_n^-(X, Y\|c)$ to denote the situation in which $F(y|x, c)$ has the following property: $n$ is the smallest integer such that, for any $y$, $i \in [1, m], j \in [n, m), i + j \leq m$,

$$F(y|x_i, c) \ FSD \ F(y|x_{i+j}, c).$$

Notice that $S_1^+(X, Y\|c)$ and $S_1^-(X, Y\|c)$ correspond to $S^+(X, Y\|c)$ and $S^-(X, Y\|c)$, respectively.

With the generalized qualitative relationship, we may extend the applicability of the reported theorems. Assume that the qualitative relationship between $Y^i$ and $Z$ is $S_n^{\sigma^i}(Y^i, Z\|e, SB(Y^i))$ rather than $S^{\sigma^i}(Y^i, Z\|e, SB(Y^i))$. To obtain bounds of $F(z|e)$, we need to replace $F(z|e, y)$ by $\hat{F}(z|e, y)$ such that $\hat{F}(z|e, y) \geq F(z|e, y)$ for all $z$ and $y$ and that $S^{\sigma^i}(Y^i, Z\|e, SB(Y^i))$ holds in the abstract network. When the values of $F(z|e, y)$ are available, we may be able to replace $F(z|e, y)$ in such a manner, and we may apply the theorems to compute bounds of $F(z|e)$.

Assume that we have $S_2^+(Z, T)$ in the network shown in following figure, and consider the task of computing bounds of $F(t|e)$. In this case, none of the reported theorems are applicable as they are stated. However, assuming that $Z$ has three possible states, we may replace $F(t|z)$ using the following method to create an approximate version of the network in which $S^+(Z, T)$ holds.

$$\hat{F}(t|z_1) = \max(F(t, z_1), F(t, z_2))$$
$$\hat{F}(t|z_2) = \max(F(t, z_1), F(t, z_2))$$
$$\hat{F}(t|z_3) = F(t|z_3)$$

As a result, we have (a) $\hat{F}(t|e, y1, y2) \geq F(t|e, y1, y2)$ for all $y1$ and $y2$, $S^+(Y1, T\|e, Y2)$ and $S^-(Y2, T\|e, Y1)$ in this approximate network, and we may apply the theorems to compute upper bounds of the the actual $F(t|e)$ using this approximate network.

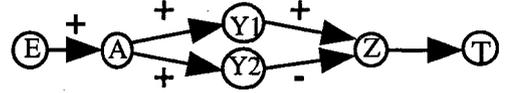

Figure 4: Application of the generalized qualitative relationships.

In addition, the generalized qualitative relationship simplifies the strengthening and weakening operations. For instance, if we have $S_n^\sigma(A, Y^i)$, the right hand sides of (2) and (3) become $\min_{j \in [l-n+1, l]} F(y|x_j, pa(Y))$ and $\max_{j \in [k, k+n-1]} F(y|x_j, pa(Y))$, respectively.

## 5 Potential Applications

We can apply the ISSA algorithm along with existing inference algorithms for QPNs (Wellman 1990; Druzdzel & Henrion 1993) for qualitative probabilistic inference. For instance, bounds of probability distributions can be used to resolve ambiguous qualitative relationships between variables in QPNs. We report applications of bounds to this task in another paper (Liu & Wellman 1998).

We may also combine the ISSA algorithm with inference algorithms with QPNs to return purely qualitative solutions and incrementally more precise numerical solutions. Assume a Bayesian network in which links are already marked with qualitative signs. After we infer qualitative relationship between variables, we may want to know something about the numerical relationship between these variables. The ISSA algorithm can compute monotonically tightening bounds of conditional CDFs.

The tightening bounds of CDFs allow us to compute tightening intervals for the expected values of variables. The expected value of a variable $Z$ is $\int_Z z \, dF(z|e)$. Therefore, by Theorem 3, the intervals for the expected value of $Z$ must tighten, if we compute the expected value using the lower and upper bounds of $F(z|e)$.

**Theorem 3 (cf. (Fishburn & Vickson 1978))** *Let $g(x)$ be a monotonically increasing function of a random variable $X$, and $F(x)$ and $F'(x)$ denote two cumulative distribution functions of $X$. Then, $F(x)$ FSD $F'(x)$ iff $\int g(x)dF(x) \geq \int g(x)dF'(x)$.*

352  Liu and WellmanThe tightening intervals for expected values can be useful for applications that use *monotone decisions* (Wellman 1990). Let the function $\delta_u(x)$ choose the value of decision variable $D$ to maximize the utility $u$ for a variable $x$.

$$\delta_u(x) = \underset{d}{\operatorname{argmax}}\ u(d, x)$$

It can be shown that $\delta_u(x)$ increases monotonically in $x$ if $u$ is a supermodular function (Ross 1983).

**Definition 4** *A function $u$ is called supermodular if, for all $d_1 \leq d_2$ and $x_1 \leq x_2$: $u(d_1, x_2) + u(d_2, x_1) \leq u(d_1, x_1) + u(d_2, x_2)$.*

A common example of supermodular functions is the utility function defined as the negative of the squared-error loss function $L(d, x) = (d - x)^2$ (cf. (Berger 1985)). The tightening intervals for the expected values of $X$ imply that the range of $D$ in which the optimal $d$ exists is decreasing, thereby helping decision makers to focus on fewer alternatives of $D$ in applications with monotone decisions.

In addition, our theorems can be used to compute the bounds of travel costs on *stochastically consistent* transportation networks defined in (Wellman, Ford, & Larson 1995). We can apply our theorems to compute bounds of travel costs, and extend the applicability of our theorems to a broader class of transportation networks, using the generalized qualitative relationships defined in Section 4.4.

## 6 Related work

Our approach is similar to many existing algorithms for computing bounds of probability values in that we ignore some information that specifies the exact distributions of Bayesian networks in the computation. For instance, *bounded conditioning* ignores some cutset instance to compute probability bounds and consider more instances to improve the bounds (Horvitz, Suermondt, & Cooper 1989). Search-based algorithms search more probable assignments of all variables and use these instances to compute probability bounds. The bounds can be improved by considering more instances that are less probable (Poole 1993). The *incremental term computation* method takes advantage of the idea of more probable instances but uses a different strategy to compute the bounds (D'Ambrosio 1993). The *localized partial evaluation* algorithm removes selected nodes from networks to compute probability intervals and recovers selected nodes to improve intervals (Draper & Hanks 1994). Our algorithm ignores some distinction of states of selected variables to compute approximations and recovers distinction of states to improve the approximations.

Similar to some other algorithms, ours assumes special numerical properties of the underlying distributions of Bayesian networks. For instance, Poole (1993) and D'Ambrosio (1993) design algorithms that are best for networks with skewed distributions. Jaakkola and Jordan (1996) develop techniques for computing bounds of likelihood for sigmoid Bayesian networks. Our algorithm requires the existence of qualitative relationships among some variables.

Some algorithms use the maximal and minimal values of a set of numbers in computing the desired bounds. For instance, the *mini-buckets* algorithm uses maximizing and minimizing functions to bound the values of functions for mini-buckets (Dechter 1997). Our algorithm is more close to Poole's method (1997), but we use bounds of conditional CDFs and Poole uses bounds of conditional probability values.

Our algorithm is different from existing algorithms in some aspects. First, we directly compute the bounds of conditional CDF of interest, i.e., $F(z|e)$. In contrast, some algorithms (Dechter 1997; Poole 1997) compute upper (lower) bounds of $\Pr(z|e)$ by dividing upper (lower) bounds of $\Pr(z, e)$ by lower (upper) bounds of $\Pr(e)$. Another difference is that we compute bounds using point-valued information rather than propagating bounds in the computation (Draper & Hanks 1994). Finally, we require the networks be specified with point-valued probabilistic information. We compute bounds of CDFs rather than exact CDFs for saving computational cost. The imprecision is an artifact of the computation algorithm. There is a school of research that works on modeling uncertain situations using probability intervals or other alternatives, and they also study algorithms for computing probability intervals (Walley 1991; Thöne, Güntzer, & Kießling 1992; Chrisman 1996; Luo *et al.* 1996; Cozman 1997).

## 7 Summary

We report an application of an extended version of our ISSA algorithm for computing lower and upper bounds of conditional cumulative distribution functions of interest. The algorithm takes advantage of qualitative relationships among variables for bounding the distributions. We show that the bounds tighten monotonically with iterations of computation, thereby providing a guarantee for improving quality of approximations for anytime computation. In particular, when used in applications with supermodular utility functions, the monotonically tightening bounds help reduce the set of decision alternatives in each iteration of the ISSA algorithm.

## References

Berger, J. O. 1985. *Statistical Decision Theory and Bayesian Analysis*. Springer-Verlag.

Boddy, M., and Dean, T. L. 1994. Deliberation scheduling